\documentclass[conference,compsoc]{IEEEtran}

\usepackage{cite}
\usepackage{graphicx}
\usepackage{float}
\usepackage{amsmath}
\usepackage{multirow}
\usepackage{rotating}
\usepackage{tikz}
\usepackage{verbatim}
\usepackage{url}

\usepackage{breakurl}

\begin{document}
\title{LoGANv2: Conditional Style-Based Logo Generation with Generative Adversarial Networks}

\author{\IEEEauthorblockN{Cedric Oeldorf}
\IEEEauthorblockA{Department of Data Science and Knowledge Engineering\\
Maastricht University\\
Maastricht, The Netherlands\\
Email: cedric.oeldorf@gmail.com}
\and
\IEEEauthorblockN{Gerasimos Spanakis}
\IEEEauthorblockA{Department of Data Science and Knowledge Engineering\\
Maastricht University\\
Maastricht, The Netherlands\\
Email: jerry.spanakis@maastrichtuniversity.nl}}

\maketitle

\begin{abstract}
Domains such as logo synthesis, in which the data has a high degree of multi-modality, still pose a challenge for generative adversarial networks (GANs). Recent research shows that progressive training (ProGAN) and mapping network extensions (StyleGAN) enable both increased training stability for higher dimensional problems and better feature separation within the embedded latent space. However, these architectures leave limited control over shaping the output of the network, which is an undesirable trait in the case of logo synthesis. This paper explores a conditional extension to the StyleGAN architecture with the aim of firstly, improving on the low resolution results of previous research and, secondly, increasing the controllability of the output through the use of synthetic class-conditions. Furthermore, methods of extracting such class conditions are explored with a focus on the human interpretability, where the challenge lies in the fact that, by nature, visual logo characteristics are hard to define. The introduced conditional style-based generator architecture is trained on the extracted class-conditions in two experiments and studied relative to the performance of an unconditional model. Results show that, whilst the unconditional model more closely matches the training distribution, high quality conditions enabled the embedding of finer details onto the latent space, leading to more diverse output. 
\end{abstract}

\section{Introduction}

Since their inception, Generative Adversarial Networks (GANs) showed the way to a whole new generation of neural network (NN) applications \cite{goodfellow2014generative}. By enabling NN-based approaches to tasks that traditionally require human-sourced creativity, we have seen many incredibly interesting uses such as generating images from text \cite{reed2016generative}, composing music \cite{yang2017midinet}, fashion design \cite{kim2017learning}, illustrating animation characters \cite{jin2017towards} and even creating emojis from peoples faces \cite{taigman2016unsupervised}. 

Tackling a creative domain such as the illustrative design process can provide creative professionals with tools that both assist and augment their work. One such domain requiring vast amounts of creativity is that of brand/logo design. With most Americans exposed to 4,000 to 20,000 advertisements a day \footnote{\url{https://www.forbes.com/sites/forbesagencycouncil/2017/08/25/finding-brand-success-in-the-digital-world/}}, companies are paying ever increasing attention to their branding. This puts pressure on designers to come up with aesthetic yet innovative and unique designs in an attempt to set their designs apart from the masses. 

GANs could assist designers by either providing them with inspiration or by reducing the number of design iterations undergone with clients. 
A problem with GAN generated content is that samples are created from an unknown noise distribution known as the input latent code. In order to facilitate specification based content generation, the user must be able to shape this latent input code in such a way that it allows for an intuitive determination of the output's style.

The implementation of class conditions, which guide the network to produce output associated with the features of a specific class, could serve as a possible approach to this. One challenge lies in the fact that such a model would require logos to be sectioned into classes based on their visual characteristics whilst logo characteristics are hard to define. 

Furthermore, the high degree of multi-modality of logos \cite{sage2018logo} increases the complexity of the task. Whilst being a problem that has been tackled in previous research \cite{sage2018logo, mino2018logan} which showed somewhat stable 32 \(\times\) 32 pixel results, such a small resolution is undesirable for real use as most details are not identifiable. Considering that model complexity exponentially increases with image size, another challenge involves generating higher resolution and more detailed logos. 

With the various aspects of the problem in mind, we engage with the following questions throughout this research paper:

\begin{itemize}
    \item How can we extract high-quality and easily definable conditions from logos?
    \item Can we increase input/output image resolution whilst retaining stable training and sensible results?
    \item How does enforcing conditions on a GAN affect performance on multi-modal data?
\end{itemize}

Both our data and python implementation are available via \texttt{GitHub}  \footnote{https://github.com/cedricoeldorf/ConditionalStyleGAN}. 
\section{Background and Related Work}


First experiments involved conditional GANs for logo synthesis and manipulation \cite{sage2018logo}. Results showed that including conditioning labels as input to a GAN has a positive effect on training stability as it promotes feature disentanglement, given that the synthetic labels are meaningful. State-of-the-art performance was achieved in terms of quantitative performance metrics by their Wasserstein GAN (WGAN) extended with an auxiliary classifier (WGAN-AC). However, they concluded that results looked more appealing in terms of human judgment when produced by their layer conditional deep convolutional GAN (DCGAN-LC). 

Shortly thereafter, notable results were achieved through the use of colour as a condition within a WGAN-AC using gradient penalty \cite{mino2018logan}. Albeit achieving promising results, the paper confirmed that the conditions need to convey meaningful information that can guide the logo in feature disentanglement, else the generator sets back to a random state.

\subsection{Progressive Training}

 The issue of GAN stability is tackled by adjusting the training methodology to start training on low resolution images and add higher resolution layers as training progresses \cite{karras2017progressive}. Considering that low resolution images hold fewer modes and less class information \cite{odena2017conditional}, the network is enabled to learn large-scale patterns in the data and then pivot from learning coarse to progressively more fine detail as resolution increases \cite{karras2017progressive}. Not only does this stabilize training, but also speeds it up through the fact that the network does not need to tackle the task of immediately mapping the intermediate latent space to high resolution imagery \cite{karras2017progressive}.

\begin{figure}[!t]
    \centering
    \includegraphics[width=3.0in]{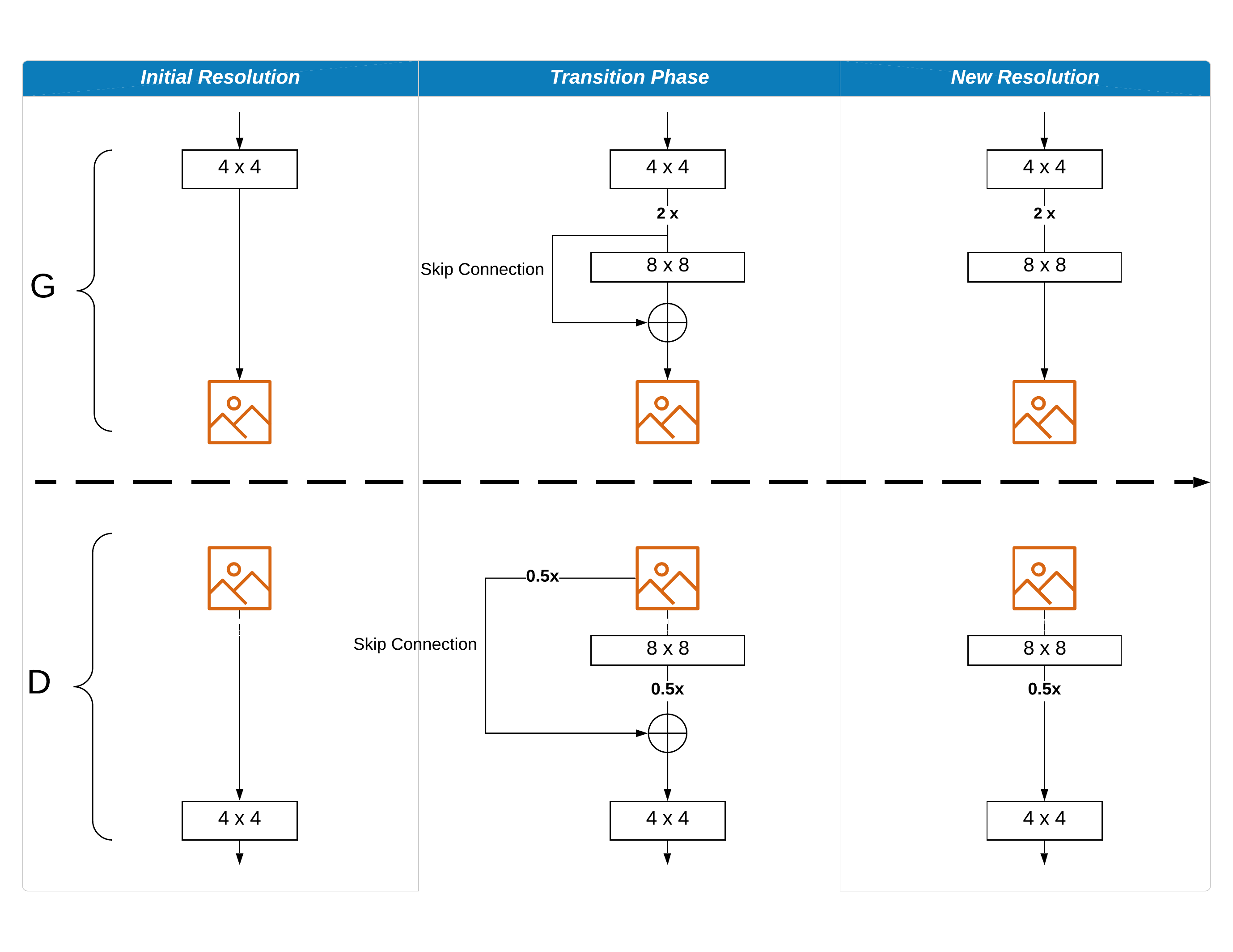}
    \caption{Progressive growing of the GAN architecture involves the addition of a higher resolution layers via a transition phase.}
    \label{fig:transition}
\end{figure}

An overview of such a methodology can be seen in Figure \ref{fig:transition}. Training of both the generator G and the discriminator D starts at a low resolution of 4 \(\times\) 4 pixels, but increases as training advances. In order to avoid a "shock" to the network when adding new layers, a transition phase outlined in the center of Figure \ref{fig:transition} is implemented. During the transition phase, skip connections are introduced to assist the pass-over to a higher resolution. After a few iterations these connections are removed and training continues. It is important to note that no layers are frozen during the process, with the entire network remaining trainable throughout the process.

Whilst progressive training solves GAN stability for high resolution training to a large extent, the embedding of a complex data distribution onto a (relatively) small latent space results in unavoidable feature entanglement. That is, multiple features are mapped to single positions in the space. Such entanglement leads to significantly difficulty in controlling the output of the network, which, if we refer to our problem statement, is something we would like to achieve. This brings us to an extension of this architecture presented in the following section.

\subsection{Style-Based Generator}

By combining progressive training, a mapping network and adaptive instance normalization, feature entanglement as mentioned in Section 2.1 can be combated with the StyleGAN architecture \cite{karras2018style}. 

\begin{figure}[!t]
    \centering
    \includegraphics[width=3.0in]{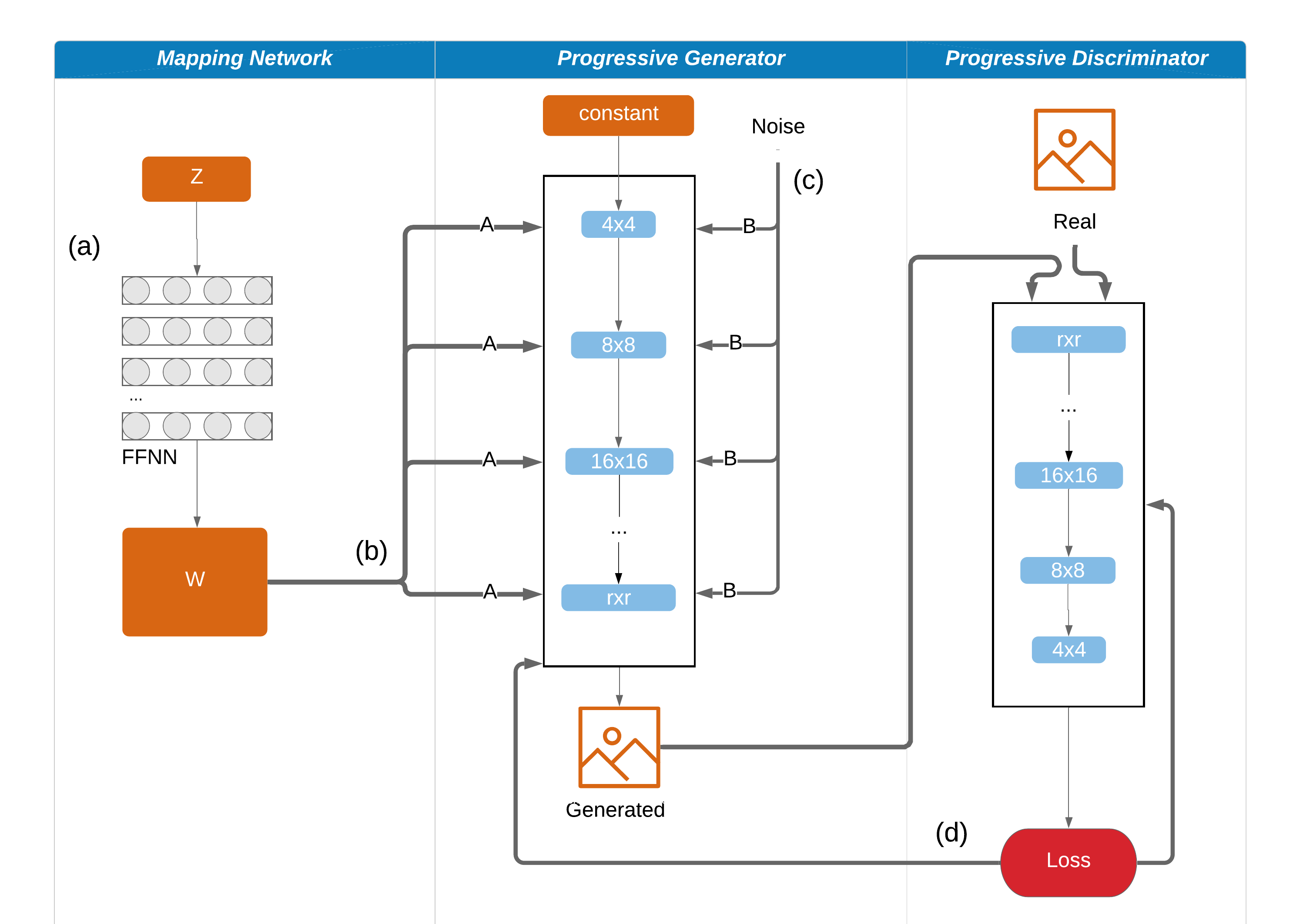}
    \caption{An outline of StyleGAN architectural elements. The mapping network transforms the initial latent vector z into w by feeding it through a feed-forward neural network (FFNN). The progressively grown generator takes both latent vector and noise inputs at all layers. The discriminator grows with the generator and returns a loss metric which is backpropagated to both networks. }
    \label{fig:sgan}
\end{figure}

We outline the upcoming subsections by exploring the alphabetical markings on Figure \ref{fig:sgan}. Marked as "(a)" in the figure, we start by describing the mapping network, which provides the input to the generator and serves as the baseline combatant to feature disentanglement. This is followed by marking "(b)", which describes how the output of the mapping network is fed into the progressive generator and "(c)" explores how stochastic variation can be introduced to the model. Marking "(d)" lies close to the loss function, which we explore for our specific case in Section 4.

\subsubsection{Mapping Network}\label{mapping_section}

In order to gain independent control over individual low- to high-level features, a feed-forward neural network can be used as a mapping network. As opposed to feeding a random vector straight into the generator, the input is projected onto an intermediate latent space \(w\) by being fed through a mapping network \cite{karras2018style}. This latent space allows controls of "styles" within convolutional layers at each resolution using Adaptive Instance Normalization explored in the next subsection .

\subsubsection{Adaptive Instance Normalization}

Traditionally in stabilization methods, each convolutional layer is accompanied by batch normalization \cite{ioffe2015batch}, which eases training complexity. However, significant improvement concerning style transfer results was observed when instance normalization (IN) was first introduced \cite{ulyanov2017improved}.

\begin{equation}
     IN(x) = \gamma (\frac{x-\mu(x)}{\sigma(x)}  ) + \beta  
     \label{IN}
\end{equation}

IN is seen in Equation \ref{IN}, where \(\mu(x)\) and \(\sigma(x)\) represent scale and bias. It normalizes input with respect to a certain style which is defined by the parameters \(\gamma\) and \(\beta\). In order to allow this equation to adapt to any given style, an adaptive affine transformation can be used. 

\begin{equation} 
    AdaIN(x,y) = \sigma(y) (\frac{x-\mu(x)}{\sigma(x)}  ) + \mu(y) 
\end{equation}

Adaptive instance normalization (AdaIN), as shown in the above equation, takes an additional input of style y in order to replace the two parameters of IN with a scale \(\sigma(y)\) and a bias \(\mu(y)\) \cite{huang2017arbitrary}.

In context of the mapping network in section \ref{mapping_section}, the intermediate space \(w\) is transformed into the scale and bias of the AdaIN equation. Noting that \(w\) reflects the style, AdaIN, through it's normalization process, defines the importance of individual parameters in the convolutional layers. Thus, through AdaIN, the feature map is translated into a visual representation.

\subsubsection{Stochastic Variation}

\begin{figure}[!t]
    \centering
    \includegraphics[width=3.0in]{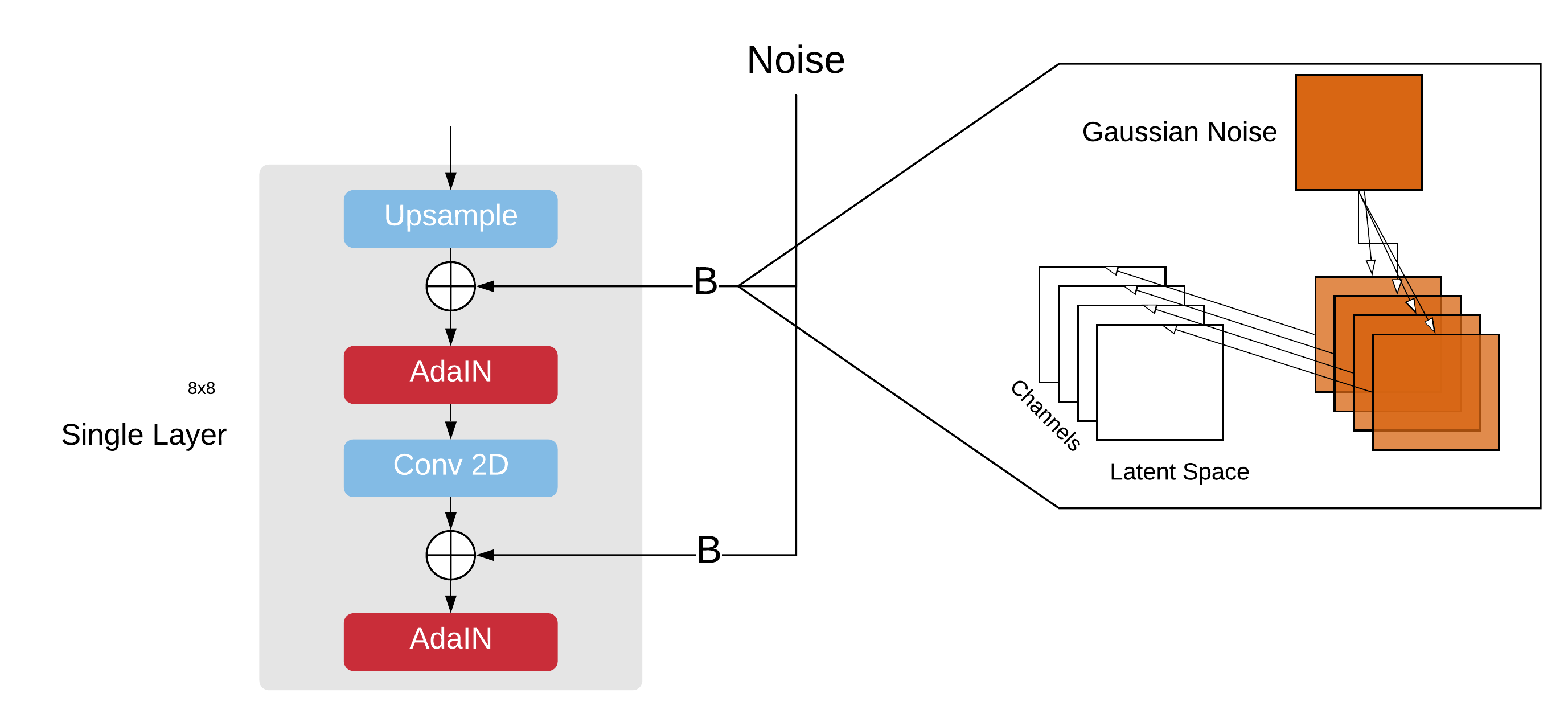}
    \caption{Inserting Noise into the Generator}
    \label{fig:noise}
\end{figure}

As features are controlled by the mapping network and AdaIN, it was found that a random noise vector is not needed as initial input, but can be replaced by a constant \cite{karras2018style}. However, in order to introduce stochastic variation into the model, which is traditionally done by inserting noise into the random input vector, it was proposed to add noise to each channel of the individual convolutional layers as seen in Figure \ref{fig:noise} \cite{karras2018style}.

The noise inputs take the form of two-dimensional matrices sampled from a Gaussian distribution. These are then scaled to match the dimensions within the layer and applied to each channel \cite{karras2018style}. This introduces variation within that feature space.

\subsubsection{Drawback: Limited control}\label{stylegandrawback}
In the scenario such as wanting to generate a specific type of logo for a client, we would want to give generator a specification by which it should synthesize a logo. Although the addition of the mapping network allows for increased control of style/features through W, this space is is quite large and likely imperfect. Each value of W would need to be mapped to a feature in order to allow for human control of the synthesized logo. 

As a goal in this paper is to be able to generate logos based by some form of input, a possible solution to this problem is introduced as the model used in this this work in Chapter 4.

\section{Data Preprocessing and Label Extraction}
This section introduces both data and methods used to extract meaningful class-conditions from logos. There are three stages, data preprocessing in which the logos are cleaned and prepared, feature extraction in which we project the logos into a quantifiable space and clustering in which K-means clustering applied to the mentioned space. 

\subsection{BoostedLLD}

The data used to train the models for our experiments is based off of the LLD-logo \cite{sage2018logo} dataset, which was preprocessed and boosted with additional images in order to be a more precise fit for our problem domain.

\subsubsection{Large Logo Data}

First introduced by Sage et al. (2018), the large logo dataset (LLD) comes in two forms, LLD-icon and LLD-logo. The icon version consists of 32px favicons, whereas the logo version consists of up to 400px logos scraped from twitter. Whilst previous research in logo synthesis \cite{sage2018logo, mino2018logan} made use of or focused on the icons set, we opted for the logo version as the higher resolution is in line with our research goals. 

Whilst the LLD-logo set consists of 122,920 logos, many of these logos consist purely of text as seen in Figure \ref{fig:textlogos}. 

\begin{figure}[h]
    \centering
    \includegraphics[width=0.2\textwidth]{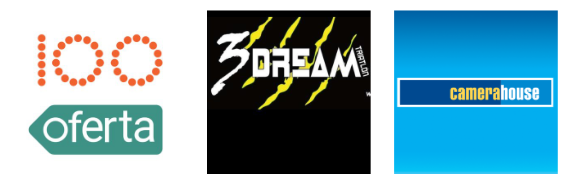}
    \caption{Examples of text-based logos}
    \label{fig:textlogos}
\end{figure}

As we would like to shift the focus of our model away from fonts and text, we decided to drop all text-based logos using the tesseract \cite{smith2007overview} open-source optical character recognition model. After dropping all images that had identifiable text on them, we were left with circa 40 000 logos.  

\subsubsection{Boosting}
\begin{figure}[h]
    \centering
    \includegraphics[width=0.2\textwidth]{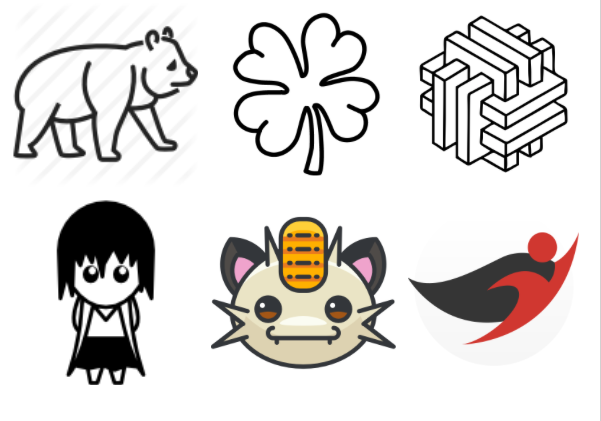}
    \caption{Examples of logos from boosting dataset}
    \label{fig:boosting}
\end{figure}

With the aim of diversifying and extending the remaining data after the preprocessing of the LLD-logo set, we scraped Google for new logo-like images. Keywords pertaining to topics such as nature, technology and illustrated characters were used in what resulted in circa 15 000 additional images for our training data. From the examples in Figure \ref{fig:boosting} we see that, whilst these images aren't official logos, they do carry the desired features of potential logos. 

\subsection{Label Extraction}
In order to train our model in a conditional manner, we need class labels for our data. These can be extracted through clustering-based methods, of which the two we used are outlined below. 

\subsubsection{Object-Classification-Based Clustering}\label{badcluster}
With one of our research goals being the extraction of human-interpretable characteristics from logos, we opted for a clustering method that makes use of Google Cloud Vision in the following three steps:

\begin{itemize}
    \item Step 1: Using the Google Vision API\footnote{https://cloud.google.com/vision/}, each logo is given 4 to 8 word labels that describe the contents. .
    
    \item Step 2: In order to bring the word labels into a quantitative space, we use a a pre-trained Word2Vec model\footnote{https://code.google.com/archive/p/word2vec/} \cite{mikolov2013efficient}. Having multiple words per logo, the midpoint between the word vectors of each individual logo is calculated and serves as the spatial representation for each logo. 
    
    \item Step 3: K-means clustering \cite{macqueen1967some} is used to partition the images into segments with the aim of having these represent unique visual properties within the logos. It is clear that, whilst there is some separation of visual characteristics, many logo styles appear in multiple clusters and it is not immediately clear what a cluster should represent. 
    
\end{itemize}{}

The poor separation of logos is what brings us to our second clustering approach in the following section. 

\subsubsection{ResNet Embedding Layer}\label{resnetquality}
Our second clustering approach follows the same steps as the first but with a different labelling technique. It makes use of the last max pooling layer of a pre-trained VGG16 \cite{simonyan2014very} network in order to project input images into a 512-dimensional feature space. Samples taken from the clusters show that certain clusters represent certain visual characteristics, more so than in section 3.2.1.

\section{Conditional Style-Based Generator}
Recalling from Section \ref{stylegandrawback}, the unconditional StyleGAN architecture only allows for limited control over the produced output. Taking a step towards lifting this limitation, we introduce multiple extensions to the StyleGAN architecture which when combined turn it into a class-conditional model.

Such conditions would not only allow the specification of a style preceding output generation, but might also assist the model in learning a larger number of complex modes \cite{sage2018logo}. Considering that many related research efforts make use of very structured image domains where images across the distribution share, for example, locality of landmarks \cite{karras2018style, arjovsky2017wasserstein}, we pay special attention to easing learning within the by nature highly multi-modal logo domain. 

\subsection{Architecture}

Motivated by the drawbacks in Section 2, we propose using a conditional framework for the StyleGAN architecture. There are two major differences between the conditional and unconditional StyleGAN architectures: (a) integrating class-conditions into intermediate latent space and (b) updating the loss function to take class-conditions into consideration.

\subsubsection{Intermediate Latent Space}

\begin{figure}[!t]
    \centering
    \includegraphics[width=3.0in]{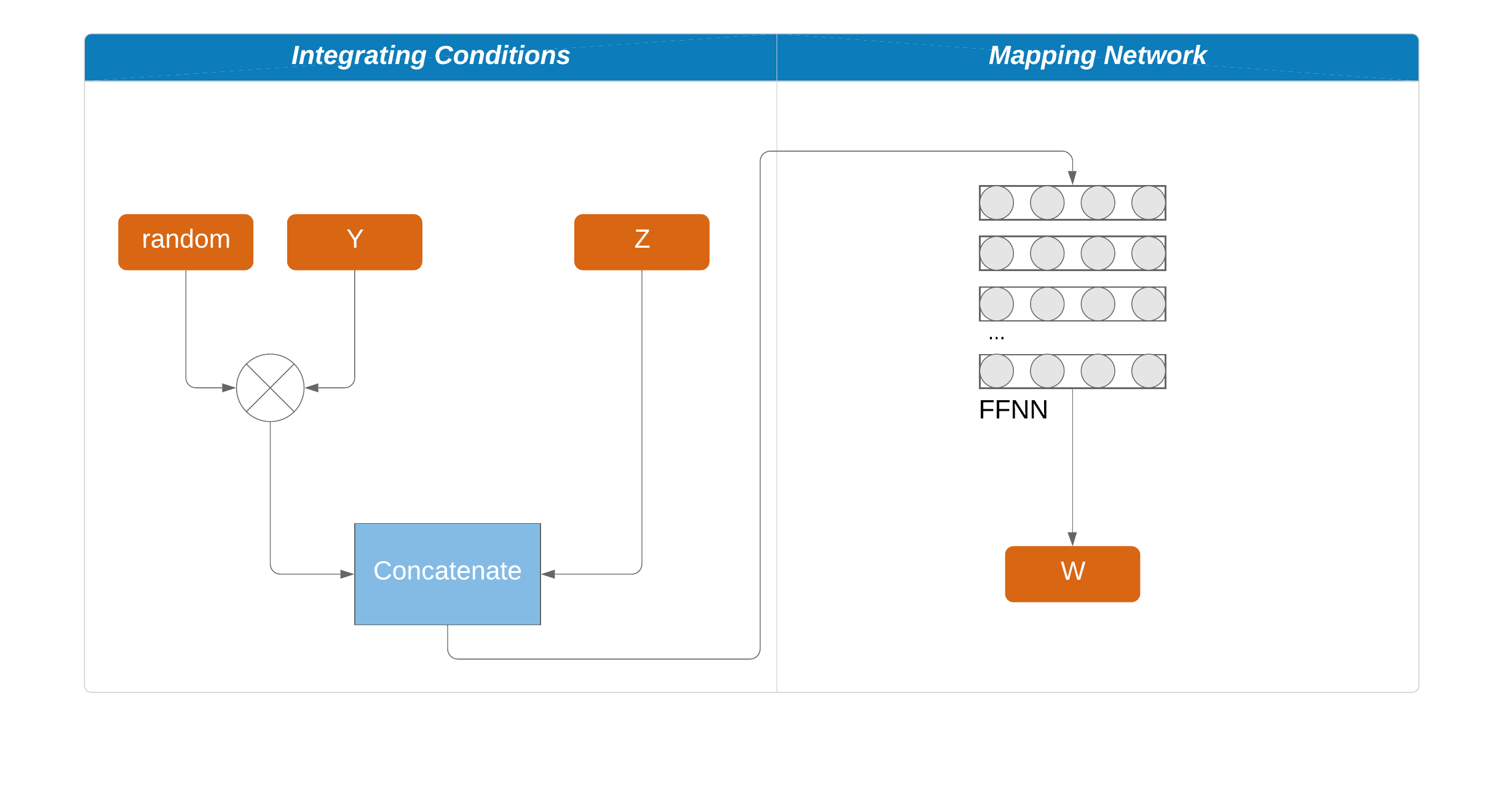}
    \caption{Producing a conditional latent vector for the StyleGAN architecture}
    \label{fig:cond}
\end{figure}

Recalling that in the StyleGAN architecture, a random vector is fed through a mapping network, which in turn produces the input vector for the generator. The incorporation of conditions into this process takes place just before the mapping network as outlined in Figure \ref{fig:cond}.

We can see how the latent embedding \(W\) depends on both the random input and the class-conditions. The process is run in mini-batches of size \(n\), which is how we define the process mathematically below. \\

We introduce two matrices in Equation \ref{matrices}. Firstly, \( Z\sim \mathcal{N}(\mu,\,\sigma^{2})\ \), which is a matrix comprised of the initial latent space of size \(n \times d \), where n is the size of the mini-batch and d is the length of the matrix. Secondly, \(Y\), which is a matrix of the one-hot-encoded conditions of size \(n \times c \), where c is the number of classes we introduce.

\begin{equation}
    Z =  \begin{bmatrix}
    l_{11} & l_{11} & \cdots & l_{1d} \\
    l_{21} & & & \\
    \vdots & & & \\
    l_{n1} & l_{n2} & \cdots & l_{nd}
  \end{bmatrix}
  ;
  Y = \begin{bmatrix}
    y_{11} & y_{11} & \cdots & y_{1c} \\
    y_{21} & & & \\
    \vdots & & & \\
    y_{n1} & y_{n2} & \cdots & y_{nc}
  \end{bmatrix}
  \label{matrices}
\end{equation}
\\
\( W\), which is the matrix holding the final latent input vectors, is calculated in Equation \ref{W}. \(Y\) is multiplied with matrix \(R \sim \mathcal{N}(\mu,\,\sigma^{2}) \) of size \(n \times c \) and concatenating the result with \(Z\). 

\begin{equation}
    W = f([Y \times R]^\frown Z)
    \label{W}
\end{equation}

The feed-forward neural network, which has also been defined as out mapping network, is represented as \( f \) in Equation \ref{W}. The result, \(W\) ends up having the same dimensions as \(L\). 

\subsubsection{Updated Loss Function}

Updating the WGAN loss function, the used loss function shows the discriminator considering the conditioning labels paired with respective output/training data \cite{isola2017image}. Equation \ref{justloss} represents this updated loss, where \(y\) are the class-conditions. 

\begin{equation}
    \bigtriangledown_{\theta D}[f_{\theta D} (x|y) - f_{\theta D}(G(w|y))] 
    \label{justloss}
\end{equation}

The architecture makes use of a gradient penalty term. Combining the term with equation \ref{justloss}, we are left with a conditional WGAN-GP loss function as seen in equation \ref{condstyleloss} below:

\begin{equation}
    \bigtriangledown_{\theta D}\underbrace{[f_{\theta D} (x|y) - f_{\theta D}(G(w|y))]}_\text{Discriminator Loss} +  \underbrace{\lambda[(|| \bigtriangledown_{\hat{x}} D(\hat{x}|y) ||_2 -1)^2]}_\text{Gradient Penalty}
    \label{condstyleloss}
\end{equation}\\

Where \( \hat{x} \) is uniformly sampled along straight lines between pairs of real and generated data points \cite{gulrajani2017improved}.

\section{Experimental Design \& Results}

The architecture undergoes three core experiments.

\begin{itemize}
    \item Experiment 1: StyleGAN architecture (no conditions)
    
    \item Experiment 2: StyleGAN conditioned on object-classification-based labels
    
    \item Experiment 3: StyleGAN conditioned on ResNet-feature-based labels
\end{itemize}

These experiments and their evaluation aim to tackle the posed research questions of whether we can generate higher resolution logos and how both the low- and high-quality conditions extracted in Section 3 affect model performance. We first describe each experiment and their respective model outputs and then go deeper into the results analysis by performing both a qualitative and quantitative evaluation in which all models are compared. The major traits we are looking for are: image quality, diversity and homogeneity within a class-condition. Image quality and diversity are explored both qualitatively and quantitatively. Homogeneity within a class-condition will be explored visually and only applies to experiment 2 and 3. 
\subsection{Quantitative Evaluation}
Using a quantitative measure for analyzing the performance of a GAN is especially difficult for a domain such as logo synthesis, as the quality of a logo is a subjective metric. 
The inception score makes use of Google's Inception classifier \cite{szegedy2015going} in order to measure the diversity of objects within generated images. If generated images hold a diverse set of recognizable objects, the inception score will be high. 

\begin{equation}
IS(G) = exp(E_{x {\sim p_{g}}} D_{K L}(P(y|x) || p(y))) 
\end{equation}

Mathematically, the score is computed as the exponential of the KL-divergence \(D_{KL}\) between distributions \(p(y|x)\) and \(p(y)\).

Whilst this was shown to correlate well with subjective human judgment of image quality \cite{ulyanov2017improved}, the scores outcome relies heavily on objects being present in the data. 

This score was extended with the proposed Frechet Inception Distance \cite{heusel2017gans}, which has been prominently used in many of the latest GAN related papers \cite{karras2017progressive,karras2018style,sage2018logo}.  The difference lies in that the distribution statistics are extracted from the feature space of an intermediate layer as opposed to from the output layer.  Mathematically it can be described as:

\begin{equation}
    FID(x,g) = || \mu_x - \mu_g||_{2}^{2} + Tr(\sum _x + \sum _g - 2(\sum _x \sum _g)^{1/2})  
\label{FIDeq}
\end{equation} 

where \(\mu\) represents the mean and \(\sum\) the covariance of the embedded layer of both \(x\), the training data, and \(g\), the generated data. We want to minimize this distance, thus the smaller FID, the higher the quality of generated images. 
The fact that the FID does not make use of object classification, makes it a good score for this paper, as logos often by nature do not contain classifiable objects. \\

Taking the elements in equation \ref{FIDeq} into account, we see the calculated score will represent both how similar the logos are to the training data in terms of quality, but also in terms of produced diversity captured by the covariance. 

\subsection{Qualitative Evaluation}

In order to subjectively measure the quality of the results, we not only view the raw results, but also analyze how well the latent embedding captured the training distribution. This is done through the truncation trick introduced below. 

\subsubsection{Truncation Trick}\label{trunc}
Within the distribution \(p(x)\) of the training data, some areas of the sampling space can be of low density due to lack of representation in the training data \cite{karras2018style}. Previous papers \cite{marchesi2017megapixel} have shown that sampling from a truncated space can improve the quality of generated results. Mathematically, this is achieved by calculating the center of mass in our latent vector \(W\) as: 

\begin{equation}
    \bar{w} = E_{z\sim P(z)}[f(z)]
\end{equation}

this point represents a type of mean generated image of the learned distribution. By increasing \( \psi \) in equation ref{truncation2}, we specify how far we would like to deviate from this center of mass. The higher the value, the more we move towards the fringes of the distribution.
\begin{equation}
    w' = \bar{w} + \psi (w - \bar{w} )
    \label{truncation2}
\end{equation}

\subsection{Experiment 1: No Conditions}
\begin{table}[!t]
\begin{tabular}{l|l}
                        & Frechet Inception Distance \\ \hline
Experiment 1 (no conditions)           & 47.4694                         \\
Experiment 2 (object label conditions)   & 71.6898                         \\
Experiment 3 (ResNet feature conditions) & 101.9211                        
\end{tabular}
\label{fig:FID}

\caption{FID scores for every experiment. A low FID score in our case implies that the synthesized logos and real logos carry closely-related visual features.}
\end{table}

\begin{figure}[!t]
    \centering
    \includegraphics[width=3.0in]{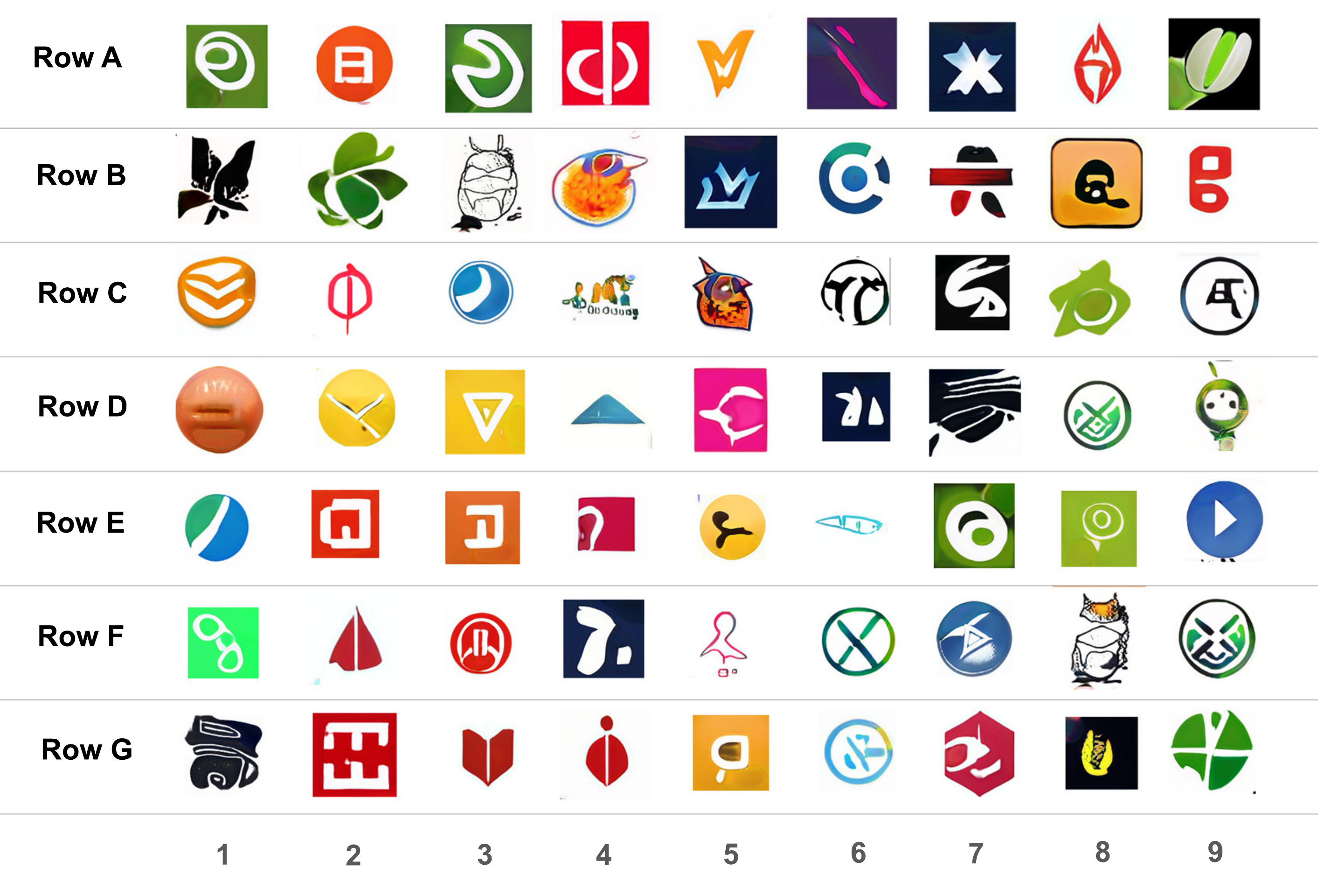}
    \caption{Experiment 1:  Synthesized logos}
    \label{fig:result_unconditional}
\end{figure}

In order to set a baseline with which to compare our conditional models with, a model was trained that does not make use of any conditions. This represents the StyleGAN architecture as it was presented in the paper it was proposed in \cite{karras2018style}. Synthesized logos are presented in Figure \ref{fig:result_unconditional}. 

\par\smallskip
\noindent\textbf{Image quality} Produced logos appear to be stable and consistently of high quality. The majority of the designs are very simple, with logos such as in B1 and B2 being rare. This could be an indication that some modes were not embedded on the latent space. According to the FID in Table \ref{fig:FID}, logos produced in experiment 1 are shown to be closest to matching the distribution of the training data.

\par\smallskip
\noindent\textbf{Diversity} We see many shape and colour repetitions with minor tweaks. However, a few truly unique designs are found such as the pattern in G1 and the character-like being in D10 of Figure \ref{fig:result_unconditional}. Additionally, when comparing this experiments behavior over different truncation values in Figure \ref{fig:trunc}, we see that the fringes of the learned embedding aren't very stable, implying that some low density modes were dropped. Based on this we can gather that output was "conservative" but in line with the models goals. 

\subsection{Experiment 2: Object Classification Based Conditions}

For this experiment, conditions based on the vectorized and clustered object classification labels were used. As was pointed out in Section \ref{badcluster}, these conditions do not show much visual separation and often embody similar styles across multiple conditions. The model produced subjectively good-looking logos as seen in Figure \ref{fig:textlogos}. However, it did suffer from slow learning, which might be indicative of low-quality conditions acting as a sort of regularizer. 

\begin{figure}[!t]
    \centering
    \includegraphics[width=3.0in]{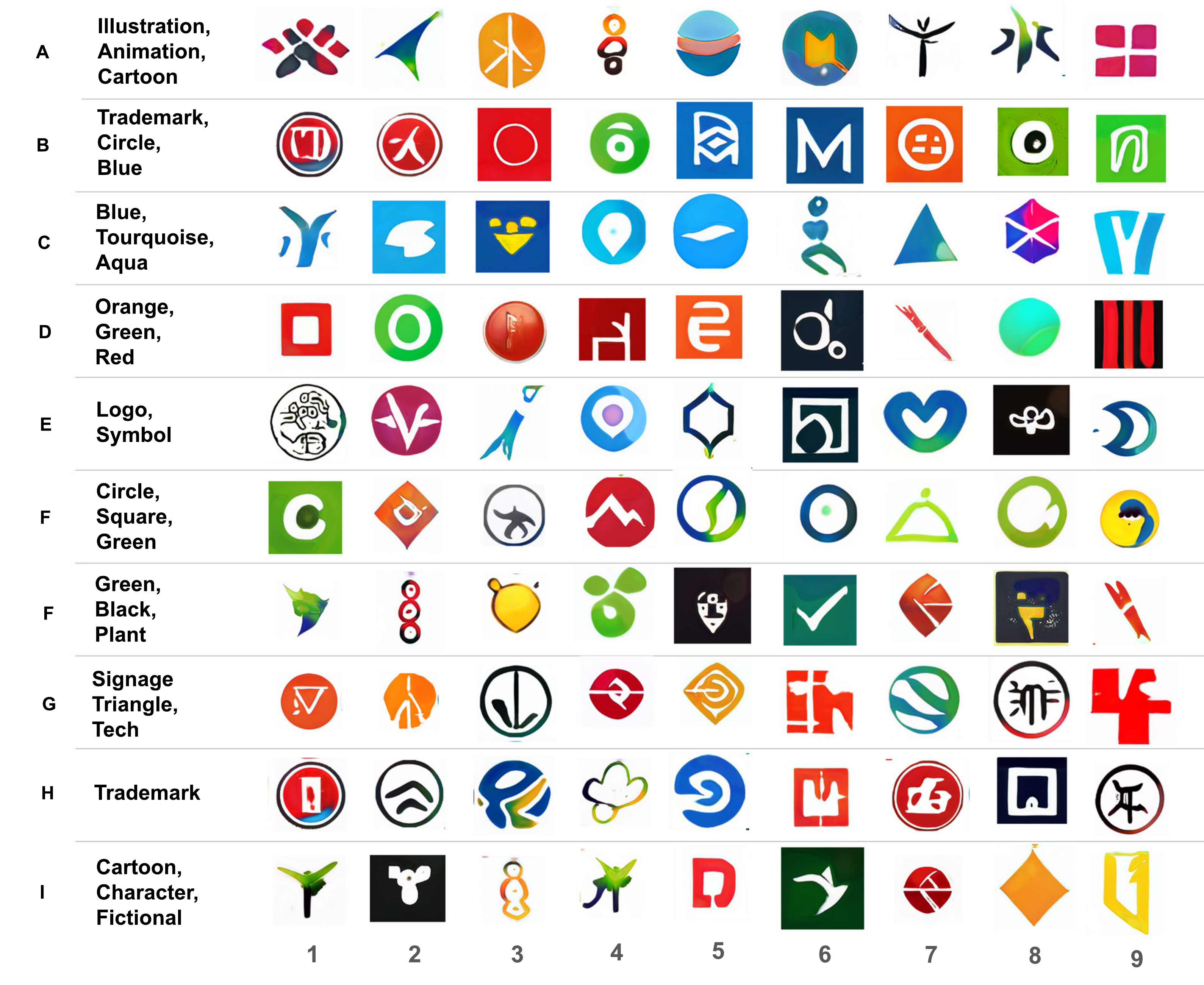}
    \caption{Experiment 2: Synthesized Logos}
    \label{fig:textlogos}
\end{figure}

\par\smallskip
\noindent\textbf{Image quality} Many examples are not as clearly defined nor as sharp as in the previous case. Occasionally, such as in B2,E6 and F7, we find logos with fine very fine details, which is indicative of the model having embedded a few complexities of the distribution. However, such fine details quickly seem to form nonsensical output such as E1, and most logos follow a general square or circle trend. Considering that the FID score in Table \ref{fig:FID} is much higher, we gather that the images do not match the training distribution well, however, through this the generated results are fairly unique.

\par\smallskip
\noindent\textbf{Diversity} Whilst we do find a lot of repetition such as in G8 and H9, the seem to generally be quite diverse compared to experiment 1. A possible reason for this is that through the conditions, the embedding of the complexities within the training data distribution becomes simpler. 

\par\smallskip
\noindent\textbf{Homogeneity}
The lack of visual separation within the conditional classes of this experiment expectantly resulted in almost no homogeneity within the conditions. We do see weak trends such as "blue" in row C, leading us to believe that with improved class conditions we would see strong visual representation of these in each row.

\subsection{Experiment 3: ResNet Feature Conditions}

\begin{figure}[!t]
    \centering
    \includegraphics[width=3.4in]{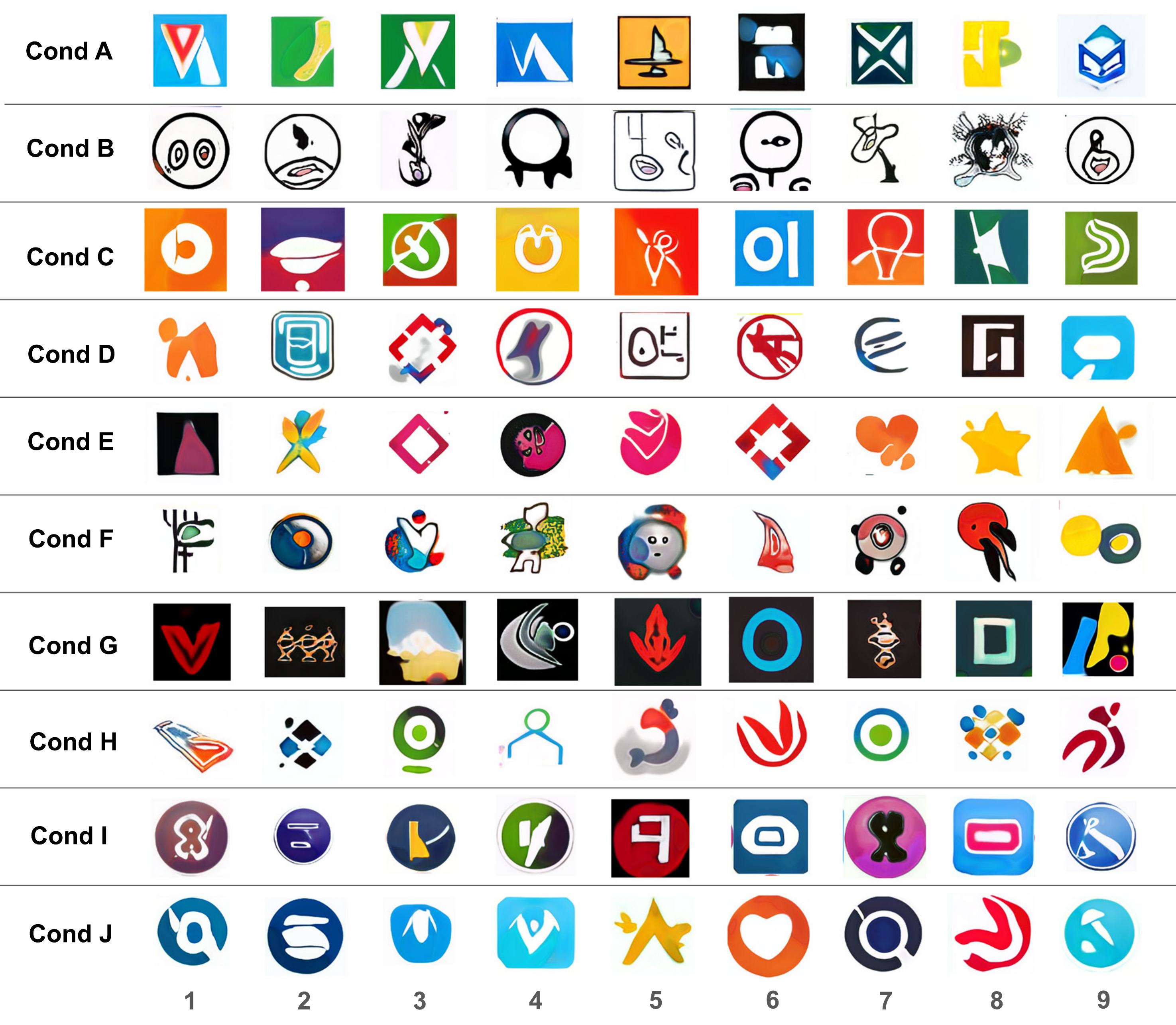}
    \caption{Experiment 3: Synthesized Logos}
    \label{fig:reslogos}
\end{figure}

The third experiment made use of conditions based on the clustered output of an embedded ResNet layer. These conditions were of significant higher quality compared to those of Experiment 2, as pointed out in Section \ref{resnetquality}. The high quality labels seem to support the models learning by feeding it meaningful information. We display synthesized logos in Figure \ref{fig:reslogos}.

\par\smallskip
\noindent\textbf{Image Quality} Whilst it is surely the most creative of the three models described, the logos produced in this experiment see both clearly defined, high-quality examples (e.g. C5,C8 or D9), but also incoherent shapes that don't seem to resemble anything we see in the training data (e.g. F7 or H1). If we take into account that the FID of experiment 3 is more than double that of experiment 1, it can be deduced that finer output detail leads to quick divergence from the training distribution. Unfortunately the FID metric cannot take the visual creativity of the output into account, a property at which in our opinion experiment 3 excelled. 

\begin{figure}[!t]
    \centering
    \includegraphics[width=3.1in]{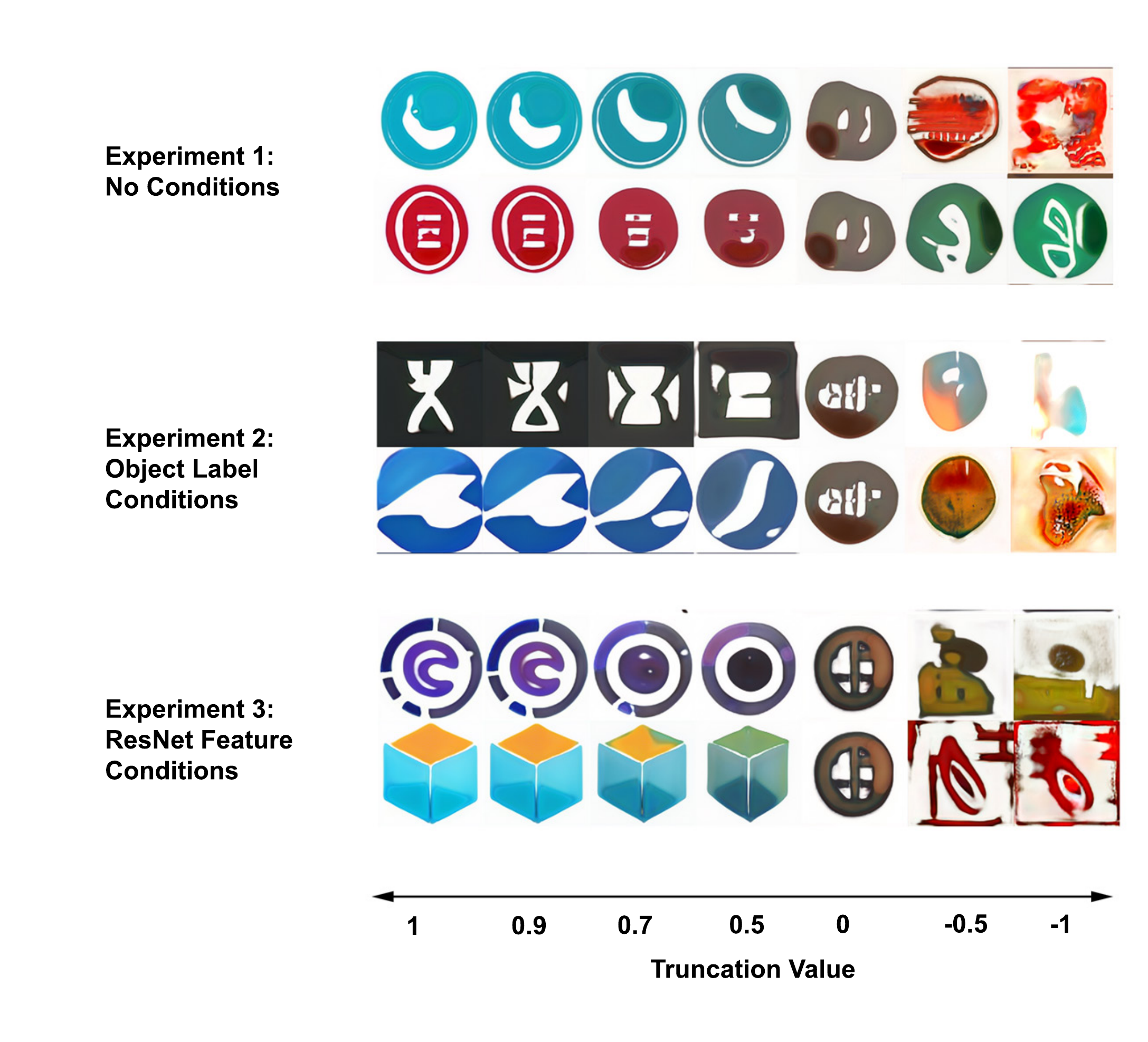}
    \caption{Experimental results over different truncation values}
    \label{fig:trunc}
\end{figure}

Furthermore, in Figure \ref{fig:trunc}, experiment 3 clearly retains the highest image quality of the three models at a truncation value of 1. This is in line with our expectations, where the high-quality conditions enabled the model to better learn the complexities of the training data distribution with respect to each class. This is furthermore confirmed by looking at a truncation value of 0, which represents the "average" logo produced by each network. Experiment 3's average holds arguably finer detail compared to the others.

\par\smallskip
\noindent\textbf{Diversity} Even within each condition, the model consistently produces a diverse set of examples. This is indicative of the model being capable of learning even the high level features of the training data distribution. 

\par\smallskip
\noindent\textbf{Homogeneity} As expected, using visually coherent class-conditions shows a high magnitude of homogeneity of style within the conditions. We derive from this that the conditions aided style separation within the latent space, allowing for the mapping of specific conditions to features commonly found within each. 

\section{Conclusion}

In this paper, we experimented with a conditional extension to the popular StyleGAN architecture in order to control the output of logo generation. We showed that meaningful conditions can help the model capture modes that might not be captured by an unconditional model. 
High-quality conditions inserted into the latent code proved to be a viable approach to controlling the output of a synthesis network. 

We found that extracting easily definable classes from logos poses a challenge. Whilst the method involving the clustered output of ResNet layer embedding provided us with visually distinct groupings, we saw within the object-classification based method that creating automated language-labels does not result in descriptions that define the visual characteristics of logos. Moreover, both conditional and unconditional StyleGAN architectures enabled stable training for a resolution 4 times larger compared to previous research. We attribute this outcome to the fact that progressive training simplifies the distributions to be embedded on the latent code. Furthermore, experiments have shown that the introduction of class-conditions in the model enables it to learn a larger number of modes in the context of highly multi-modal data. It became apparent that the meaningful separation of data into classes eases learning for the model. Having been fed high-quality information, the model is faster to shift its focus onto high level features, resulting in much more detailed logo synthesis. Lastly, a trade-off between the robustness and detail of generated output was observed. The conditional models were shown to produce more unique logos with fine features, but at the cost of also producing some nonsensical output. The unconditional model however, which generated mostly very simple but realistic logos, did not see the same amount of nonsensical output. We additionally proposed a new data set that shifts the focus of the LLD collection away from text-based designs and more towards illustrations. Although the size of the data set was significantly reduced, we believe that the higher quality played a significant factor in the success of our models.

In future, it would be desirable to head towards a text-input based architecture, especially when considering the use-cases of the logo domain. Keeping our approach in mind, a possible scenario might include replacing the class-conditions input with word vector representations. However, whilst the challenge of producing meaningful word labels from logos still goes unsolved, first word-based logo generation experiments could be run using easy-to-identify characteristics such as colour and shape. If successful, this would offer a very simple and straightforward way of controlling the latent embedding, and with that the architecture's output.

\bibliographystyle{IEEEtran}
\bibliography{references}{}

\end{document}